\documentclass[conference]{IEEEtran}
\usepackage{cite}
\usepackage{amsmath,amssymb,amsfonts}
\usepackage{algorithmic}
\usepackage{graphicx}
\usepackage{textcomp}
\usepackage{multirow}
\usepackage[table,xcdraw]{xcolor}
\def\BibTeX{{\rm B\kern-.05em{\sc i\kern-.025em b}\kern-.08em
    T\kern-.1667em\lower.7ex\hbox{E}\kern-.125emX}}
\usepackage{booktabs}
\usepackage{subcaption}
\begin{document}

\title{Evaluating the Fairness of the MIMIC-IV Dataset and a Baseline Algorithm: \\ Application to the ICU Length of Stay Prediction}

\author{%
  Alexandra Kakadiaris \\
  Department of Data Science \\
  Columbia University \\
  New York, NY 10027 \\
  \texttt{ak5087@columbia.edu} \\
}

\maketitle
\thispagestyle{plain}
\pagestyle{plain}

\begin{abstract}

  This paper uses the MIMIC-IV dataset to examine the fairness and bias in an XGBoost binary classification model predicting the Intensive Care Unit (ICU) length of stay (LOS). Highlighting the critical role of the ICU in managing critically ill patients, the study addresses the growing strain on ICU capacity. It emphasizes the significance of LOS prediction for resource allocation. The research reveals class imbalances in the dataset across demographic attributes and employs data preprocessing and feature extraction. While the XGBoost model performs well overall, disparities across race and insurance attributes reflect the need for tailored assessments and continuous monitoring. The paper concludes with recommendations for fairness-aware machine learning techniques for mitigating biases and the need for collaborative efforts among healthcare professionals and data scientists.
  
\end{abstract}

\section{Introduction}

The Intensive Care Unit (ICU) provides critically ill patients with intensive and continuous medical care, typically involving advanced monitoring, life support systems, and specialized medical personnel to help prevent the progression of severe illness \cite{weil2011intensive}. The growing need for critical care in patients with severe health conditions strains ICU capacity. This strain manifests in limited bed availability, increased workloads for medical and hospital staff, and consequent delays in ICU admissions, ultimately contributing to higher morbidity and mortality rates \cite{hempel2023prediction}. 

The ICU's patient length of stay (LOS) is a process metric commonly employed to indicate ICU efficiency and effectiveness \cite{devos2007quality}. Extended ICU stays correlate with elevated care costs and resource consumption, whereas premature discharges may result in medical complications, heightened ICU readmission risks, and increased mortality \cite{alghatani2021predicting}. The significance of predicting ICU length of stay lies in its pivotal role in resource allocation, discharge planning, and overall patient management. Understanding the factors contributing to variations in length of stay can empower healthcare providers to make informed decisions and optimize the utilization of resources. LOS explains 85\% to 90\% of the variation in hospital costs between patients \cite{rapoport2003length}.

However, it is imperative to recognize that the effectiveness of these predictive models may be compromised if considerations of fairness and equity are overlooked. The impact of these models on patient outcomes and resource allocation may vary across different demographic and socioeconomic groups, potentially perpetuating health disparities. Ensuring fairness in these models and the data these models use for training is crucial for promoting equitable healthcare practices and avoiding unintended consequences that could disproportionately affect specific patient populations. Addressing these concerns is vital for ethically deploying predictive models in the ICU setting.

\subsection{Clinical/Machine Learning Motivation}

In recent years, the healthcare landscape has witnessed a surge in data-driven approaches to optimize patient care and resource allocation \cite{mathews2015conceptual}. Among the critical areas of interest is the ICU length of stay prediction, a pivotal metric for healthcare providers to anticipate resource needs and enhance patient outcomes \cite{stone2022systematic}.
Fairness has various definitions in different domains of health care. Specifically, fairness addresses whether an algorithm treats sub-populations equitably. The issue of fairness has recently attracted more attention as ML-driven decision support systems are increasingly applied to practical applications \cite{li2022improving}. Algorithmic unfairness, or bias, may introduce or exaggerate health disparities in health care \cite{paulus2020predictably}. Furthermore, the exploration of fairness in healthcare, specifically in the context of insurance types, addresses broader concerns about disparities in access to quality care and the potential impact on patient outcomes. This paper will treat fairness as an analysis of whether the model makes equitable decisions for sensitive groups. In our paper, the sensitive attributes are race and insurance. 

\subsection{Related Work and Contribution} 
Research has demonstrated that many nonclinical factors significantly affect health care, including race, sex, education, socioeconomic status, and insurance, which have
been shown to alter access to and quality of medical
care, leading to decreased resource utilization \cite{englum2016association}. There is a lack of academic literature comprehensively leveraging MIMIC-IV data, specifically by categorizing the data based on insurance types. This gap in research hinders our understanding of potential discrepancies in algorithmic prediction length of stay across different insurance categories.
If the model exhibits disparities in predicting the length of stay based on insurance types, it could perpetuate biases and discriminatory practices. Such biases might lead to unequal distribution of resources, potentially disadvantaging individuals with specific insurance coverage. Therefore, an accurate estimate of the patient's LOS in the ICU, based on the patient's initial health data, helps healthcare management in the appropriate resource allocation and better planning for the future \cite{hempel2023prediction}.

The contribution of this paper is evaluating the performance of an XGBoost binary classification model and its training data for predicting the Intensive Care Unit (ICU) length of stay (LOS) across different sensitive attributes, explicitly focusing on insurance types and race. The objective is to assess potential disparities and biases in the model's predictions and highlight the importance of fairness in healthcare predictive modeling. The paper emphasizes the need for a nuanced evaluation of the model's performance within specific subgroups or sensitive groups, such as Medicare and Medicaid patients or different racial categories. The findings underscore the importance of tailoring predictive models to account for variations in healthcare outcomes among distinct demographic groups and the need for ongoing refinement to ensure fairness and equity in healthcare practices.

\subsection{Objectives}
This research project evaluates an XGBoost binary classification strategy by defining two classes with four days, the mean average length of stay of patients over 24 hours, as a threshold. The first class predicts that patients' stay will be {\it short} ($<$ four days), and the second class predicts that patients will be {\it extended} ($>=$ four days). This algorithm emphasizes explicitly exploring bias and fairness. The goal is to implement a baseline algorithm tailored for ICU LOS prediction, focusing on observing potential bias towards sensitive groups. Following this, we conduct a detailed evaluation of the accuracy achieved by this algorithm, paying close attention to any disparities in predictive performance across different demographic groups. As a crucial aspect of our investigation, we emphasize a comprehensive assessment of potential sample bias within the training data set. To ensure fairness in our predictions, we undertake an in-depth analysis of the baseline model, emphasizing the importance of unbiased outcomes for all individuals, irrespective of the sensitive group. Our analysis delves into the fairness of results for every sensitive group, providing valuable insights into the equitable performance of the ICU LOS prediction model to assess whether the predictive model is free from biases that may disproportionately affect certain sensitive groups. This analysis aims to analyze these biases in the data and discuss how we should address the fairness questions if a model performs differently across sensitive groups.

\section{Methods}
This section outlines the data set and the algorithms for predicting the ICU LOS. The publicly accessible MIMIC-IV data set is the foundation for our experiments, wherein a defined cohort is established with specific features tailored for the analysis. Data collected within the initial 24 hours of ICU admission was collected, categorizing features of interest into four groups: demographics, routine vital signs obtained upon ICU admission, laboratory measurements, and medication information. These measurements were obtained from the first 24 hours of ICU admission. Data cleaning and preparation precede the analysis phase, culminating in evaluating the model's performance.

{\em Dataset:}
The Medical Information Mart for Intensive Care (MIMIC) IV (version 2.1) database, a publicly available database encompassing information on patients admitted to Beth Israel Deaconess Medical Center, Boston, MA, from 2008 to 2019 \cite{johnson2022mimic} was used. The information available includes patient measurements, orders, diagnoses, procedures, treatments, and de-identified free-text clinical notes \cite{johnson2022mimic}.

{\em Pre-Processing:}
A series of inclusion and exclusion criteria were used to construct the cohort for predicting ICU LOS. First, all ICU stays longer than 24 hours were included without restricting whether a single patient had multiple visits. That means that one patient might have multiple ICU stays. This approach aims to capture diverse patient experiences within the ICU setting. ICU stays that resulted in patient mortality were also retained to address potential censoring. Treating death as a censoring event allows us to capture and analyze the full spectrum of patient outcomes. For patients meeting the inclusion criteria in the ICU stays, we mapped routine vital signs, explicitly saving the first routine vital signs collected for each patient at each ICU stay. This ensures a representative and standardized set of physiological data for subsequent analysis. Then, information on each patient's laboratory results, procedures, and medications was extracted within 24 hours of their ICU stay. This time frame is a crucial window for capturing initial clinical interventions and the patient's health profile during the early stages of critical care.

{\em Feature Extraction:}
Our pipeline begins with four data frames — patients, labs, chart events (vital signs), and pharm (medications administered) — each containing numerous variables totaling hundreds or thousands for each patient. In the approach, the selective inclusion of specific variables was deliberate to mitigate the risk of introducing bias into our analysis. By incorporating all available variables from the patients, labs, events, and pharm data frames, we aimed to maintain a comprehensive and unbiased representation of the underlying health data. The decision to include all variables is rooted in acknowledging that singling out particular variables for analysis may inadvertently introduce selection bias, skewing the results towards the chosen variables and potentially overlooking critical insights in the broader data set.

Two issues arose with including every measurement that appeared in all data sets. The first issue was that many patients did not have information on most attributes related to vital signs and labs. To solve this issue, it was decided to run the models (i) by not imputing, (ii) by imputing using the mean, and (iii) by imputing using the median of the values of that specific lab. Throughout this paper, the analysis is presented for each of these choices. Imputing with the median and median for NA values in the database is a pragmatic strategy to prevent the ML model from misinterpreting missing data as abnormal instances, especially in healthcare, where specific tests or medications might not be administered if not clinically indicated. Creating a model with these three variations ensures a more accurate representation of the data set, mitigating potential biases and enabling the model to better capture the underlying patterns in the data.

{\em Baseline Model:}
We employed the XGBoost algorithm for our predictive modeling due to its robust performance in handling numeric data and capturing complex patterns in large datasets. The features considered in the model were restricted to numeric values, explicitly focusing on laboratory test results, vital sign measurements, and binary indicators representing whether a patient took medication during the first 24 hours in the ICU. The focus was on effectively capturing the quantitative aspects of patient health without including metrics, such as race, insurance, and gender, that could bias the data. 

{\em Implementation Details:}
Applying the inclusion and exclusion criteria and the preprocessing steps (e.g., data cleaning and normalization) were executed using Python programming in a Jupyter Notebook environment. Data manipulation and analysis libraries such as Pandas and NumPy were utilized to streamline the preprocessing pipeline. The ICU length of stay predictive models were implemented within the same Jupyter Notebook environment. It is used to train and evaluate the XGBoost algorithm using the sci-kit-learn and TensorFlow libraries. All computations, including data preprocessing and model training, were performed on a virtual machine (VM) instance hosted on the Google Cloud Platform (GCP). The VM instance on the Google Cloud Platform had specifications tailored to the computational demands of the study. The choice of machine type, storage, and memory resources was made to optimize the efficiency of the data processing and machine learning tasks. The VM instance provided the necessary computational resources to handle the sizeable MIMIC-IV dataset and execute resource-intensive machine-learning tasks. The scalability and reliability of the Google Cloud Platform contributed to the efficiency of the computational workflow.

The Jupyter Notebook is available on GitHub (https://github.com/akakadiaris/FairnessofMIMICIV) to encourage transparency, reproducibility, and further collaboration within the research community.

\section{Results}

\subsection{Sensitive groups imbalance in MIMIC-IV and the training data set}
Before we begin the analysis of our data set, it is crucial to acknowledge and address class imbalances across various demographic attributes. Examining the data set reveals disparities that warrant careful consideration to ensure equitable and unbiased model outcomes.

The dataset exhibits class imbalances across various demographic attributes. In terms of race, the whole dataset is imbalanced, with a predominant representation of the White population (67.17\%). At the same time, the Black-African American, Hispanic, Asian, and other racial categories have comparatively lower proportions. Similarly, gender distribution is slightly imbalanced, with 52.90\% females and 47.10\% males. Notably, the insurance attribute demonstrates imbalance, particularly in the Medicaid, Medicare, and Other categories, with Medicare being the most prevalent (33.33\%).

In the training data, class imbalance persists across demographic attributes. In terms of race, there is an imbalance, with 68.34\% of the data representing the White population, while Black, Hispanic, Asian, and other racial categories have lower proportions. The gender distribution remains slightly imbalanced, with 52.90\% females and 47.10\% males. Notably, the insurance attribute shows imbalances, with Medicare being the most prevalent (46.19\%), followed by Other (46.60\%), and a representation from Medicaid (7.21\%).

Comparing the whole dataset to the training data, some variations are evident. The training data exhibits a slightly higher representation of the White population (68.34\%) compared to the whole dataset (67.17\%) and a notable decrease in the representation of the Multiple Race category (0.34\% in training data vs. 0.13\% in the whole dataset). There is a considerable shift in insurance, with Medicare being more dominant in the training data (46.19\%) compared to the whole data set (33.33\%). 

\subsection{Analysis of Bias in the Training Data}
\begin{figure}[htb]
  \centering
  \begin{subfigure}{0.48\linewidth}
    \includegraphics[width=\linewidth]{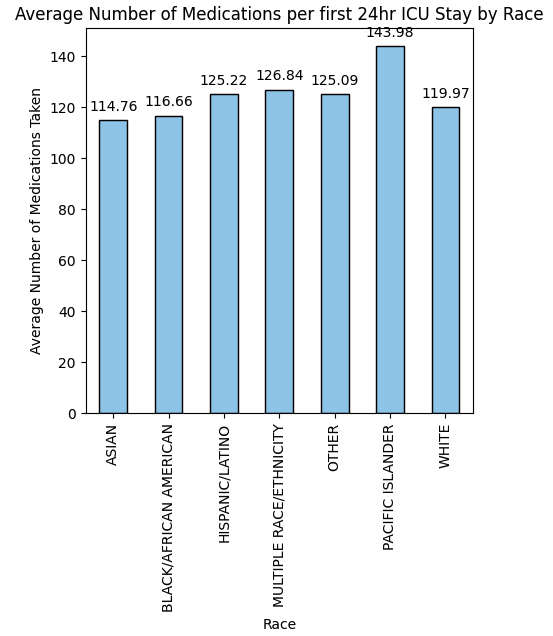}
    \caption{}
    \label{fig:screenshot4}
  \end{subfigure}%
  \hfill
    \hfill  
    \hfill
  \begin{subfigure}{0.45\linewidth}
    \includegraphics[width=\linewidth]{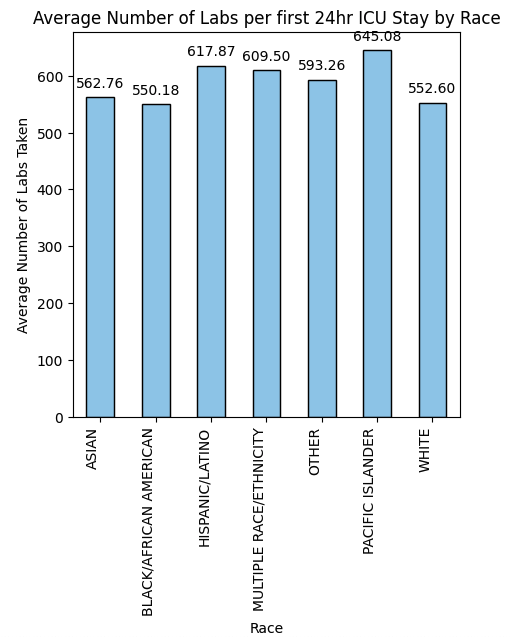}
    \caption{}
    \label{fig:screenshot5}
  \end{subfigure}%
  \hfill
    \hfill
      \hfill
  \begin{subfigure}{0.50\linewidth}
    \includegraphics[width=\linewidth]{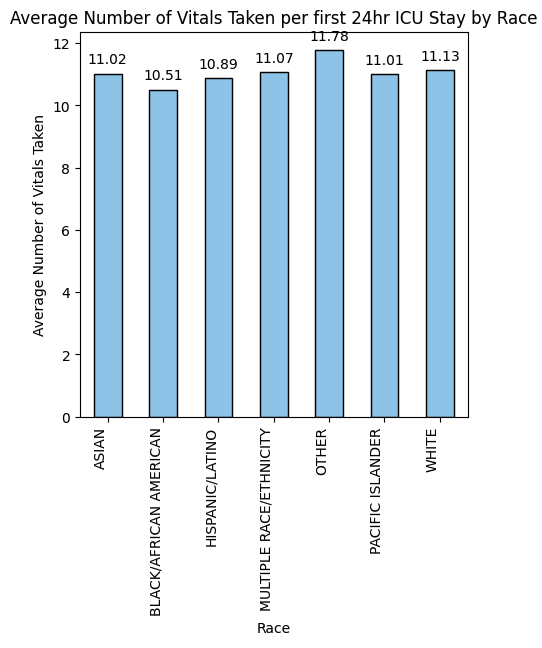}
    \caption{}
    \label{fig:screenshot6}
  \end{subfigure}
  \caption{Depiction of Race breakdown. (a) Average number of medications per ICU stay, (b) average number of labs per ICU stay, and (c) average number of vitals per ICU stay.}
  \label{fig:multi_screenshots2}
\end{figure}

\begin{figure}[htb]
  \centering
  \begin{subfigure}{0.55\linewidth}
    \includegraphics[width=\linewidth]{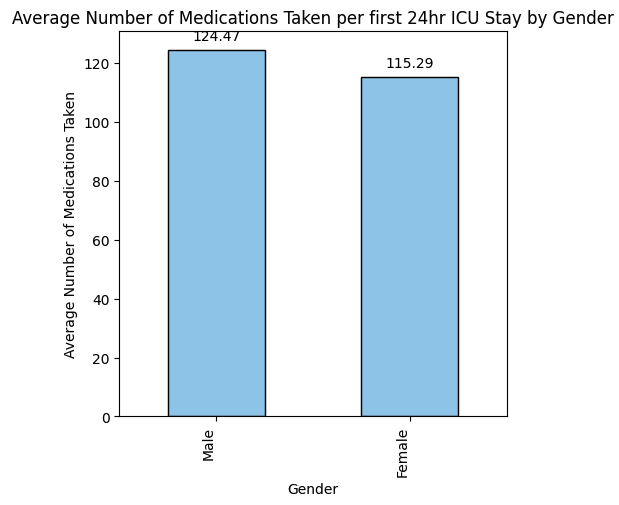}
    \caption{}
    \label{fig:screenshot7}
  \end{subfigure}%
  \hfill
  \begin{subfigure}{0.45\linewidth}
    \includegraphics[width=\linewidth]{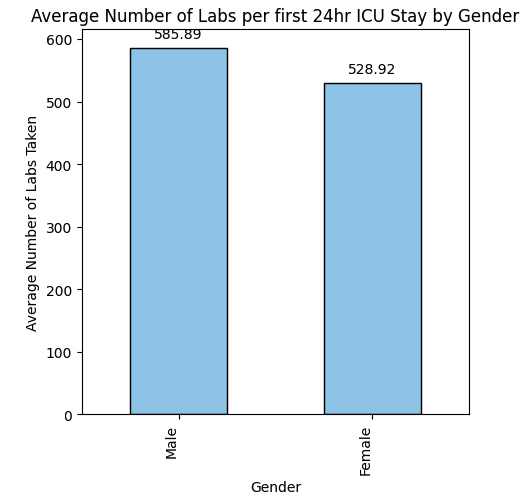}
    \caption{}
    \label{fig:screenshot8}
  \end{subfigure}%
  \hfill
   \hfill
    \hfill
  \begin{subfigure}{0.50\linewidth}
    \includegraphics[width=\linewidth]{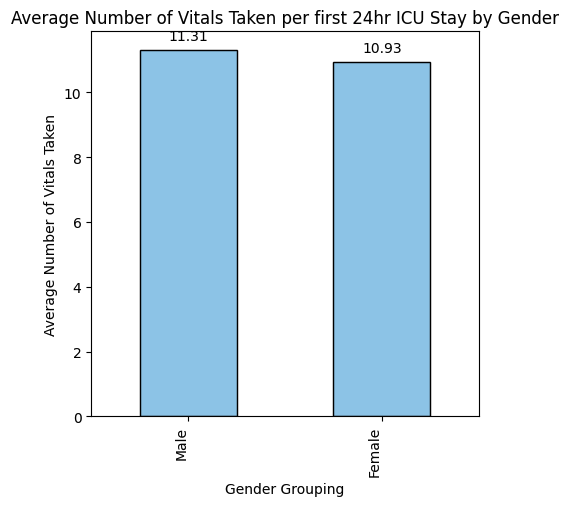}
    \caption{}
    \label{fig:screenshot9}
  \end{subfigure}
  \caption{Depiction of gender breakdown. (a) The average number of medications per ICU stay, (b) the average number of labs per ICU stay, and (c) the average number of vitals per ICU stay.}
  \label{fig:multi_screenshots3}
\end{figure}

\begin{figure}[htb]
  \centering
  \begin{subfigure}{0.55\linewidth}
    \includegraphics[width=\linewidth]{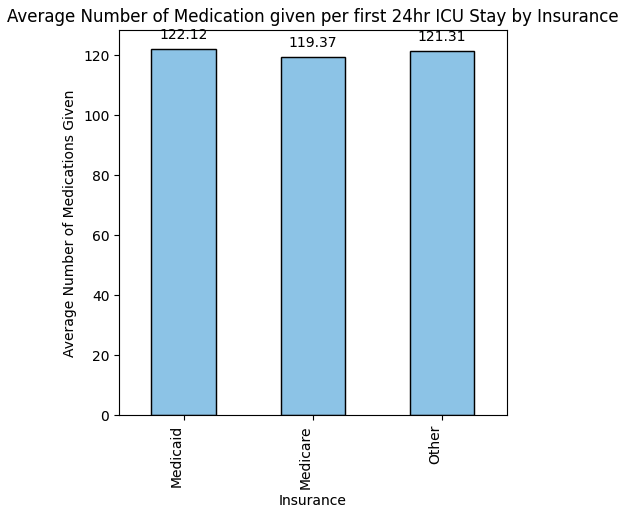}
    \label{fig:screenshot1}
    \caption{}
  \end{subfigure}%
  \hfill
  \begin{subfigure}{0.45\linewidth}
    \includegraphics[width=\linewidth]{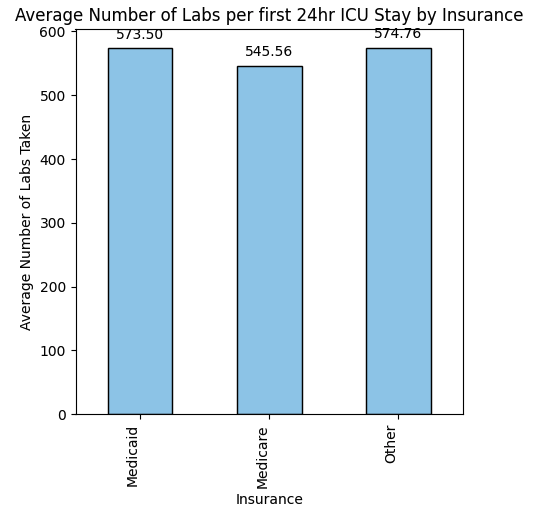}
    \label{fig:screenshot2}
    \caption{}
  \end{subfigure}%
  \hfill
  \begin{subfigure}{0.5\linewidth}
    \includegraphics[width=\linewidth]{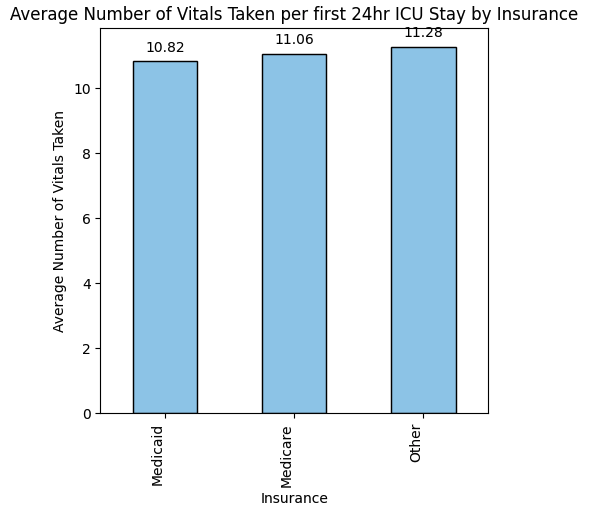}
    \label{fig:screenshot3}
    \caption{}
  \end{subfigure}
  \caption{Depiction of insurance breakdown. (a) Average number of medications per ICU stay, (b) average number of labs per ICU stay, and (c) average number of vitals per ICU stay.}
  \label{fig:multi_screenshots1}
\end{figure}

In assessing present biases within the dataset, a comprehensive examination focused on key healthcare metrics for each sensitive attribute related to race, gender, and insurance. Specifically, the average number of labs, vital signs, and medications administered in the first 24 hours of admission were scrutinized to discern potential disparities in healthcare experiences among different demographic groups. This was conducted to uncover nuanced patterns and potential biases. Figures \ref{fig:multi_screenshots2}, \ref{fig:multi_screenshots3} and \ref{fig:multi_screenshots1} examine potential biases. These figures reveal that Black/African American patients appear to undergo fewer laboratory tests and are prescribed fewer medications compared to other demographic groups.
Conversely, Pacific Islander patients exhibit a trend of undergoing a higher number of labs and taking more medications during the initial ICU period. That said, there seems to be no significant variation between other races. These also suggest that female patients tend to experience a higher average number of laboratory tests per day while being prescribed fewer medications compared to other gender groups.

\begin{figure}[ht]
  \centering
  \includegraphics[width=0.99\textwidth]{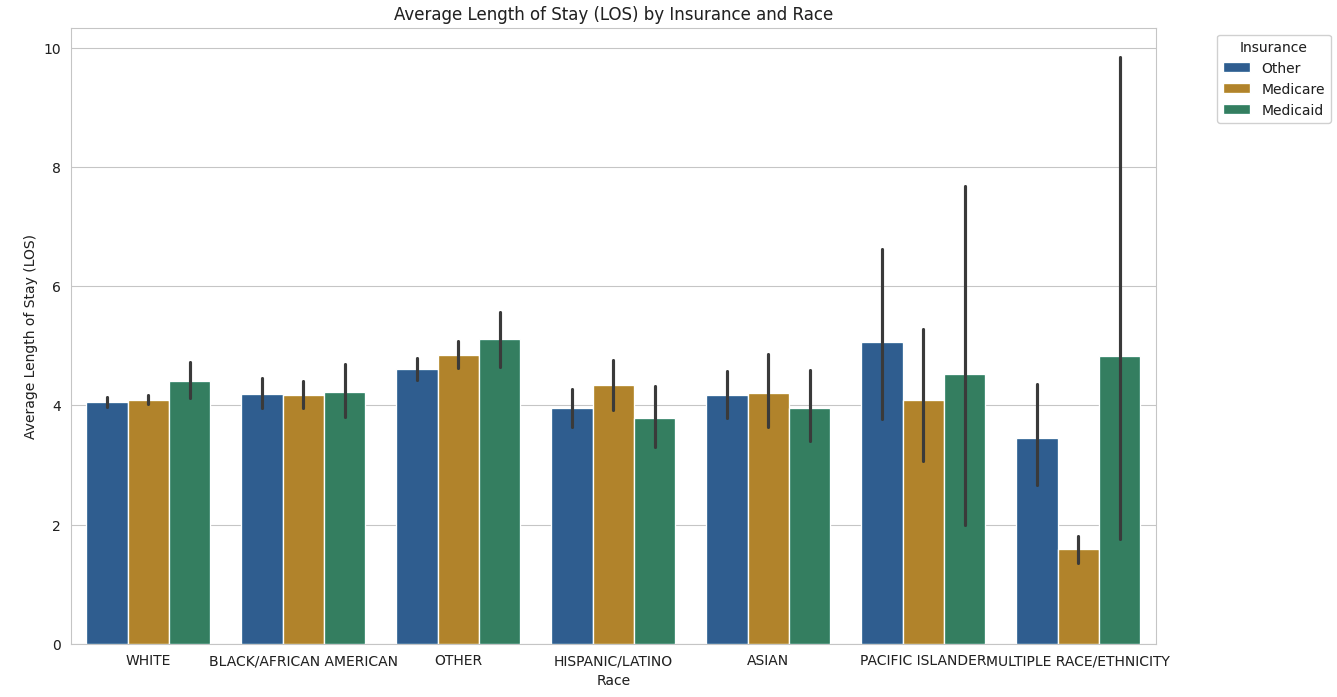}
  \caption{Length of ICU Stay by Insurance and Race}
  \label{fig:race-insurance-combo}
\end{figure}

Figure \ref{fig:multi_screenshots1} indicates that Medicare patients undergo fewer laboratory tests and are prescribed fewer medications than other insurance groups. But, as observed with race, there does not seem to be a significant difference between the three insurance types regarding these metrics. Figure \ref{fig:race-insurance-combo}
depicts the average LOS for insurance and race in one visualization. Of note is the substantial dispersion in the length of stay (number) notably observed among Pacific Islanders and multiple race patients, potentially shedding light on the considerable variation in the length of stay across different insurance categories. Besides that, while there seem to be variations in the average length of stay, one particular insurance type does not seem to affect the LOS for race. 

An ANOVA analysis, whose results are summarized in Table \ref{tab:anova_results}, was conducted to explore further and quantify the observed variations within metrics collected in the first 24 hours among sensitive groups, allowing for a comprehensive examination of potential statistically significant differences in the average number of labs and prescribed medications within the first 24 hours in the ICU. Based on these results, all of these factors have a high F-statistic that indicates a significant difference in mean length of stay between insurance types. The p-values for all factors are below the 0.05 threshold, suggesting statistical significance. There is evidence to reject the null hypothesis, indicating significant differences in the mean length of stay based on insurance type.

\subsection{Baseline Model Performance}
{\em Metrics:} The dataset was split into training and testing sets (80/20 split), with the training set used to train the XGBoost model and the testing set employed to assess its predictive performance. Doing the split taking care of imbalance and no patients same at training and testing is left to future work. We utilized standard metrics such as accuracy, precision, recall, and F1 score to evaluate the model's performance. It is important to note that the model's interpretability and generalization capabilities depend on the selected features and the dataset's representativeness. Careful consideration of potential biases and limitations is essential for interpreting the model's predictions in the context of healthcare disparities.

{\em Performance:} 
In evaluating the three predictive models' performance, the performance of these models is very similar, with accuracy of 83\% and 82\%. Overall, the area under the ROC curve (AUC-ROC) and Precision-Recall curve (AUC-PR) further substantiates the models's discriminative power, with scores for each model being around the same value. These findings are summarized in Table \ref{table:overall_metrics}. The confusion matrices (shown in Figures \ref{table:no_imputation},\ref{mean_imputation}, and \ref{median_imputation}) reveal some differences between predictions but not any significant difference in performance. 
About classification, the performance is summarized in Tables \ref{classification-no},\ref{classification-mean}, and \ref{classification-median}. The precision, recall, and F1 scores perform similarly, with only a 0.01 difference.

Despite the uniform performance observed across the three models in terms of precision, recall, and F1 score, it becomes apparent in subsequent sections of the paper that their effectiveness varies significantly when considering sensitive groups. The subsequent analysis reveals nuanced disparities in performance metrics, shedding light on the models' divergent abilities to handle instances related to different sensitive groups. This emphasizes the importance of scrutinizing model performance across various sensitive groups to unveil potential biases or discrepancies that might not be apparent when examining global performance metrics alone.

\subsection{Evaluating the Model wrt Sensitive Groups}

Transitioning to evaluating the model's performance across sensitive groups, it is crucial to delve into the nuanced implications of the results within sensitive groups. The subsequent analysis will provide insights into how the predictive model performs for distinct populations, shedding light on potential disparities and highlighting the need for nuanced considerations in healthcare outcomes prediction within these sensitive groups.

{\em Race:}
The metrics concerning race with (i) no imputation, (ii) mean imputation, and (iii) median imputation—are presented in Tables \ref{race-classification-no}, \ref{race-classification-mean}, and \ref{race-classification-median}, respectively. While precision remains consistently high for the "short" classification across different racial groups, indicating a low false positive rate for predicting the "Short" class, there are notable variations in recall and F1 score metrics, particularly for the "extended" classification.
Examining the {\em No Imputation} approach, there is a substantial difference in recall for the "extended" class between White (0.57) and Black/African American (0.57) groups, suggesting potential disparities in sensitivity. This discrepancy persists across the imputation dataset, indicating a consistent pattern of performance differences in predicting the "extended" class among racial categories. The variations underscore the need for a nuanced examination of model performance, especially concerning predicting outcomes within specific racial categories.

{\em Insurance:}
The metrics concerning insurance type with (i) no imputation, (ii) mean imputation, and (iii) median imputation are presented in Tables \ref{insurance-classification-no},\ref{insurance-classification-mean}, and \ref{insurance-classification-median}, respectively. In evaluating the model's performance across insurance categories, distinctive patterns emerge with notable variations in precision, recall, and F1 scores for predicting short and extended lengths of stay. For Medicare patients, the model demonstrates a precision of around 0.81 for short stays and 0.76 for extended stays. These numbers are lower for the model being built using training data imputed using the median. There is a higher recall for short stays (0.92) than extended stays (0.55), but again, these numbers are lower for the model being built using training data inputted using the median. Medicaid exhibits robust predictive precision for both short (0.89) and extended (0.83) stays, yet with a more considerable recall for short stays (0.95) than extended stays (0.69). Other insurance categories showcase similar trends, emphasizing the need for a comprehensive assessment of the model's predictive accuracy and sensitivity within specific insurance groups. These findings underscore the importance of tailoring predictive models to accommodate variations in healthcare outcomes across diverse insurance classes.

\section{Discussion}
The observed disparities in the model's performance across different sensitive attributes, such as insurance and race, warrant careful consideration regarding the potential impact on fairness and equitable healthcare outcomes. Notably, variations in precision, recall, and F1 scores for predicting short and extended lengths of stay highlight the nuanced challenges in developing a universally applicable predictive model. The higher precision for short stays among Medicare patients suggests a more accurate prediction for this group. In contrast, the lower recall for extended stays indicates potential underestimation in identifying extended hospitalization. This trade-off in precision and recall necessitates a delicate balance when implementing the model in real-world healthcare. Evaluating the XGBoost binary classification model for predicting ICU LOS across different demographic attributes has provided valuable insights into the model's performance. Despite achieving commendable metrics, such as accuracy and AUC-ROC, the model exhibits nuanced disparities across sensitive groups, indicating potential fairness and bias concerns. Bias focuses on systematic errors in the predictive model's outcomes, whereas fairness focuses on ensuring equitable and unbiased treatment of diverse populations by identifying these biases. 

{\em Race-Based Disparities:}
Evaluation of the model across racial groups reveals consistent disparities in predicting the "extended" class. The variation in recall for the "extended" class between White and Black/African American groups suggests potential differences. This raises concerns about the model's ability to accurately identify patients with an extended ICU stay, especially among specific racial categories. Addressing these disparities requires a nuanced approach, including potential model recalibration and increased focus on feature importance within different racial groups.

{\em Insurance-Based Disparities:} 
Analyzing the model's performance across insurance categories reveals notable predictive precision and recall variations. Mainly, Medicare patients show lower precision for "short" and "extended" stays, emphasizing the need for further refinement to enhance predictive accuracy for this group. Conversely, Medicaid exhibits robust predictive precision, suggesting that the model performs relatively well in predicting ICU LOS for this insurance category. The disparities across insurance types underscore the importance of tailoring predictive models to account for variations in healthcare outcomes among distinct insurance groups.


\subsection{Future Work}
{\em Addressing Disparities and Improving Fairness:} Identifying and addressing disparities in model performance is crucial for ensuring equitable healthcare outcomes. Implementing fairness-aware machine learning techniques, such as reweighing the training data \cite{mingyang} or adjusting prediction thresholds based on sensitive groups \cite{hardt}, might help mitigate biases. Further exploration of feature importance within specific demographic groups may unveil underlying factors contributing to performance variations, allowing for targeted improvements. Continuous monitoring and refinement of the model's performance across diverse subpopulations are essential to upholding fairness in predictive healthcare models.

{\em Data Collection and Representation:} Ensuring representative datasets for all demographic groups is fundamental for building fair and unbiased predictive models. Efforts should be directed toward collecting comprehensive and diverse healthcare data to eliminate imbalances that may contribute to disparities in model performance. Increased diversity in training data can enhance the model's ability to generalize across various demographic attributes.

{\em Model Interpretability:} Enhancing the interpretability of predictive models can aid in understanding the decision-making process and potential biases. Utilizing interpretable models or model-agnostic interpretability techniques can provide insights into the features driving predictions and help identify and rectify biases within the model.

{\em Transparency and Collaboration:} Transparency in model development and deployment is paramount. Detailed documentation of model architecture, training processes, and evaluation metrics promotes accountability and facilitates external scrutiny. Collaborative efforts among healthcare professionals, data scientists, and policymakers are essential to iteratively refine predictive models, ensuring that they align with ethical standards and promote equitable healthcare practices.

\section{Conclusion}

In conclusion, evaluating the XGBoost model for predicting ICU length of stay highlights its strengths and potential limitations, especially concerning disparities across demographic attributes. Addressing these disparities requires a multifaceted approach, including model recalibration, transparency, and stakeholder collaboration. The iterative refinement of predictive models is essential to ensure ongoing fairness and equity in healthcare outcomes. The insights gained from this study contribute to the growing body of knowledge on the intersection of machine learning and healthcare equity, guiding future efforts to develop more robust and unbiased predictive models.

\bibliographystyle{plain}
\bibliography{references}
\clearpage
\onecolumn
\section{Supplementary Material}

\begin{table}[h]
  \centering
  \caption{Summary of ANOVA Results}
  \label{tab:anova_results}
  \begin{tabular}{lrr}
  \toprule
    \textbf{Factor} & \textbf{F-statistic} & \textbf{p-value} \\ \hline
     Race & 7.1024 & $3.15 \times 10^{-31}$ \\ 
     Gender & 10.0864 & 0.0015 \\
     Insurance & 4.6511 & 0.0096 \\\hline
  \end{tabular}
\end{table}

\begin{table}[h]
\centering
\caption{Model Metrics on Testing Data}
\begin{tabular}{lccc} 
\toprule
\textbf{Metric} & \textbf{No Imputation} & \textbf{Mean Imputation} & \textbf{Median Imputation} \\ 
\midrule
Accuracy & 83\% & 83\% & 82\% \\
AUC-ROC & 0.86 & 0.85 & 0.85 \\
AUC-PR & 0.75 & 0.75 & 0.74 \\
\midrule
\end{tabular}
\label{table:overall_metrics}
\end{table}
\clearpage

\begin{table}[h]
\centering
\caption{Confusion Matrix: No Imputation}
\renewcommand{\arraystretch}{1.5}
\begin{tabular}{ll|l|r|}
\multicolumn{2}{c}{} & \multicolumn{2}{c}{Predicted} \\
\cline{3-4}
& & {\rotatebox[origin=c]{0}{Short}} & {\rotatebox[origin=c]{0}{Extended}} \\
\cline{3-4}
\multirow{0.5}{*}{{\rotatebox[origin=c]{90}{Actual}}} & Short & 5,958 & 542 \\ \cline{3-4}
& Extended & 1,010 & 1,646 \\ \cline{3-4}
\end{tabular}
\label{table:no_imputation}
\end{table}

\begin{table}[h]
\centering
\caption{Confusion Matrix: Mean Imputation}
\renewcommand{\arraystretch}{1.5}
\begin{tabular}{ll|l|r|}
\multicolumn{2}{c}{} & \multicolumn{2}{c}{Predicted} \\
\cline{3-4}
& & {\rotatebox[origin=c]{0}{Short}} & {\rotatebox[origin=c]{0}{Extended}} \\
\cline{3-4}
\multirow{0.5}{*}{{\rotatebox[origin=c]{90}{Actual}}} & Short & 5,957 & 543 \\ \cline{3-4}
& Extended & 1,040 & 1,616 \\ \cline{3-4}
\end{tabular}
\label{mean_imputation}
\end{table}

\begin{table}[h]
\centering
\caption{Confusion Matrix: Median Imputation}
\renewcommand{\arraystretch}{1.5}
\begin{tabular}{ll|l|r|}
\multicolumn{2}{c}{} & \multicolumn{2}{c}{Predicted} \\
\cline{3-4}
& & {\rotatebox[origin=c]{0}{Short}} & {\rotatebox[origin=c]{0}{Extended}} \\
\cline{3-4}
\multirow{0.5}{*}{{\rotatebox[origin=c]{90}{Actual}}} & Short & 5,950 & 550 \\ \cline{3-4}
& Extended & 1,064 & 1,592 \\ \cline{3-4}
\end{tabular}
\label{median_imputation}
\end{table}

\clearpage

\begin{table}[htbp]
\centering
\caption{Classification Report for Baseline: No Imputation}
\label{classification-no}
\begin{tabular}{lcccccc}
\toprule
    & Precision & Recall & F1-Score & Support \\
    \midrule
    0 & 0.86 & 0.92 & 0.88 & 6,500 \\
    1 & 0.75 & 0.62 & 0.68 & 2,656 \\
    \midrule
    Accuracy & & & 0.83 & 9,156 \\
    Macro Avg & 0.80 & 0.77 & 0.78 & 9,156 \\
    Weighted Avg & 0.83 & 0.83 & 0.83 & 9,156 \\
    \bottomrule
\end{tabular}
\end{table}

\begin{table}[htbp]
\centering
\caption{Classification Report for Baseline: Mean Imputation}
\label{classification-mean}
\begin{tabular}{lcccccc}
\toprule
    & Precision & Recall & F1-Score & Support \\
    \midrule
    0 & 0.85 & 0.92 & 0.88 & 6,500 \\
    1 & 0.75 & 0.61 & 0.67 & 2,656 \\
    \midrule
    Accuracy & & & 0.83 & 9156 \\
    Macro Avg & 0.80 & 0.76 & 0.78 & 9,156 \\
    Weighted Avg & 0.82 & 0.83 & 0.82 & 9,156 \\
    \bottomrule
\end{tabular}
\end{table}

\begin{table}[htbp]
\centering
\caption{Classification Report for Baseline: Median Imputation}
\label{classification-median}
\begin{tabular}{lcccccc}
\toprule
    & Precision & Recall & F1-Score & Support \\
    \midrule
    0 & 0.85 & 0.92 & 0.88 & 6,500 \\
    1 & 0.74 & 0.60 & 0.66 & 2,656 \\
    \midrule
    Accuracy & & & 0.82 & 9,156 \\
    Macro Avg & 0.80 & 0.76 & 0.77 & 9,156 \\
    Weighted Avg & 0.82 & 0.82 & 0.82 & 9,156 \\
    \bottomrule
\end{tabular}
\end{table}

\begin{table}[htbp]
\centering
\caption{Metrics per Race Classification: No Imputation}
\label{race-classification-no}
\begin{tabular}{lccccr}
\toprule
\textbf{Race} & \textbf{Classification} & \textbf{Precision} & \textbf{Recall} & \textbf{F1 Score} & \textbf{Support} \\
\midrule
White & Short & 0.84 & 0.93 & 0.88 & 884 \\
& Extended & 0.77 & 0.57 & 0.66 & 370 \\
\midrule
Black/African American & Short & 0.81 & 0.92 & 0.86 & 120 \\
& Extended & 0.77 & 0.57 & 0.65 & 60 \\
\midrule
Other & Short & 0.79 & 0.87 & 0.83 & 165 \\
& Extended & 0.72 & 0.61 & 0.66 & 94 \\
\midrule
Asian & Short & 0.86 & 0.82 & 0.84 & 45 \\
& Extended & 0.47 & 0.54 & 0.50 & 13 \\
\midrule
Hispanic/Latino & Short & 0.86 & 0.97 & 0.91 & 58 \\
& Extended & 0.80 & 0.47 & 0.59 & 17 \\
\midrule
Multiple & Short & 1.00 & 1.00 & 1.00 & 4 \\
& Extended & - & - & - & - \\
\midrule
Pacific Islander & Short & 0.50 & 1.00 & 0.67 & 1 \\
& Extended & 1.00 & 1.00 & 1.00 & 1 \\
\bottomrule
\end{tabular}
\end{table}

\begin{table}[htbp]
\centering
\caption{Metrics per Race Classification: Mean Imputation}
\label{race-classification-mean}
\begin{tabular}{lccccr}
\toprule
\textbf{Race} & \textbf{Classification} & \textbf{Precision} & \textbf{Recall} & \textbf{F1 Score} & \textbf{Support} \\
\midrule
White & Short & 0.83 & 0.93 & 0.83 & 884 \\
& Extended & 0.76 & 0.55 & 0.64 & 370 \\
\midrule
Black/African American & Short & 0.78 & 0.90 & 0.83 & 120 \\
& Extended & 0.71 & 0.48 & 0.57 & 60 \\
\midrule
Other & Short & 0.82 & 0.88 & 0.85 & 165 \\
& Extended & 0.76 & 0.65 & 0.70 & 94 \\
\midrule
Asian & Short & 0.82 & 0.82 & 0.82 & 45 \\
& Extended & 0.38 & 0.38 & 0.38 & 13 \\
\midrule
Hispanic/Latino & Short & 0.86 & 0.93 & 0.89 & 58 \\
& Extended & 0.67 & 0.47 & 0.55 & 17 \\
\midrule
Multiple & Short & 1.00 & 1.00 & 1.00 & 4 \\
& Extended & - & - & - & - \\
\midrule
Pacific Islander & Short & 0.50 & 1.00 & 0.67 & 1 \\
& Extended & 0.00 & 0.00 & 0.00 & 1 \\
\bottomrule
\end{tabular}
\end{table}

\begin{table}[htbp]
\centering
\caption{Metrics per Race Classification: Median Imputation}
\label{race-classification-median}
\begin{tabular}{lccccr}
\toprule
\textbf{Race} & \textbf{Classification} & \textbf{Precision} & \textbf{Recall} & \textbf{F1 Score} & \textbf{Support} \\
\midrule
White & Short & 0.83 & 0.92 & 0.87 & 884 \\
& Extended & 0.74 & 0.53 & 0.62 & 370 \\
\midrule
Black/African American & Short & 0.81 & 0.92 & 0.86 & 120 \\
& Extended & 0.78 & 0.58 & 0.67 & 60 \\
\midrule
Other & Short & 0.82 & 0.89 & 0.85 & 165 \\
& Extended & 0.77 & 0.65 & 0.71 & 94 \\
\midrule
Asian & Short & 0.82 & 0.82 & 0.82 & 45 \\
& Extended & 0.38 & 0.38 & 0.38 & 13 \\
\midrule
Hispanic/Latino & Short & 0.85 & 0.86 & 0.85 & 58 \\
& Extended & 0.50 & 0.47 & 0.48 & 17 \\
\midrule
Multiple & Short & 1.00 & 1.00 & 1.00 & 4 \\
& Extended & - & - & - & - \\
\midrule
Pacific Islander & Short & 0.50 & 1.00 & 0.67 & 1 \\
& Extended & 0.00 & 0.00 & 0.00 & 1 \\
\bottomrule
\end{tabular}
\end{table}

\begin{table}[htbp]
\centering
\caption{Metrics per Insurance Classification: No Imputation}
\label{insurance-classification-no}
\begin{tabular}{lccccr}
\toprule
\textbf{Insurance} & \textbf{Classification} & \textbf{Precision} & \textbf{Recall} & \textbf{F1 Score} & \textbf{Support} \\
\midrule
Medicare & Short & 0.81 & 0.92 & 0.86 & 574 \\
& Extended & 0.76 & 0.55 & 0.64 & 271 \\
\midrule
Medicaid & Short & 0.89 & 0.95 & 0.92 & 95 \\
& Extended & 0.83 & 0.69 & 0.75 & 35 \\
\midrule
Other & Short & 0.84 & 0.92 & 0.88 & 608 \\
& Extended & 0.74 & 0.58 & 0.65 & 249 \\
\end{tabular}
\end{table}

\begin{table}[htbp]
\centering
\caption{Metrics per Insurance Classification: Mean Imputation}
\label{insurance-classification-mean}
\begin{tabular}{lccccr}
\toprule
\textbf{Insurance} & \textbf{Classification} & \textbf{Precision} & \textbf{Recall} & \textbf{F1 Score} & \textbf{Support} \\
\midrule
Medicare & Short & 0.81 & 0.92 & 0.86 & 574 \\
& Extended & 0.76 & 0.54 & 0.63 & 271 \\
\midrule
Medicaid & Short & 0.89 & 0.95 & 0.92 & 95 \\
& Extended & 0.83 & 0.69 & 0.75 & 35 \\
\midrule
Other & Short & 0.83 & 0.91 & 0.87 & 608 \\
& Extended & 0.72 & 0.55 & 0.63 & 249 \\
\end{tabular}
\end{table}

\begin{table}[htbp]
\centering
\caption{Metrics per Insurance Classification: Median Imputation}
\label{insurance-classification-median}
\begin{tabular}{lccccr}
\toprule
\multicolumn{6}{c}{\textbf{Insurance Classification Metrics: Median Imputation}} \\
\midrule
\textbf{Insurance} & \textbf{Classification} & \textbf{Precision} & \textbf{Recall} & \textbf{F1 Score} & \textbf{Support} \\
\midrule
Medicare & Short & 0.79 & 0.90 & 0.84 & 574 \\
& Extended & 0.70 & 0.50 & 0.58 & 271 \\
\midrule
Medicaid & Short & 0.88 & 0.94 & 0.91 & 95 \\
& Extended & 0.79 & 0.66 & 0.72 & 35 \\
\midrule
Other & Short & 0.85 & 0.94 & 0.88 & 608 \\
& Extended & 0.75 & 0.66 & 0.66 & 249 \\
\end{tabular}
\end{table}


\end{document}